\documentclass[a4paper,twoside]{article}
\usepackage{epsfig}
\usepackage{subcaption}
\usepackage{calc}
\usepackage{amssymb}
\usepackage{amstext}
\usepackage{amsmath}
\usepackage{amsthm}
\usepackage{multicol}
\usepackage{pslatex}
\usepackage{apalike}
\usepackage{SCITEPRESS}     % Please add other packages that you may need BEFORE the SCITEPRESS.sty package.
\usepackage{adjustbox}

\usepackage{xcolor}
\usepackage{verbatim}
\usepackage{amsmath}
\usepackage{booktabs}
\usepackage{tikz}
\usepackage{hhline}
\usepackage{environ}
\makeatletter
\newsavebox{\measure@tikzpicture}
\NewEnviron{scaletikzpicturetowidth}[1]{%
  \def\tikz@width{#1}%
  \def\tikzscale{1}\begin{lrbox}{\measure@tikzpicture}%
  \BODY
  \end{lrbox}%
  \pgfmathparse{#1/\wd\measure@tikzpicture}%
  \edef\tikzscale{\pgfmathresult}%
  \BODY
}
\makeatother

\usepackage{pgfplots}
\def\checkmark{\tikz\fill[scale=0.4](0,.35) -- (.25,0) -- (1,.7) -- (.25,.15) -- cycle;}

\begin{document}

%\title{Transformer based Session Dropout Prediction for Mobile Education}
\title{Deep Attentive Study Session Dropout Prediction\\in Mobile Learning Environment}

\author{\authorname{
Youngnam Lee\sup{1},
Dongmin Shin\sup{1},
HyunBin Loh\sup{1},
Jaemin Lee\sup{1},
Piljae Chae\sup{1},
Junghyun Cho\sup{1}, \\
Seoyon Park\sup{1},
Jinhwan Lee\sup{1},
Jineon Baek\sup{1,3},
Byungsoo Kim\sup{1},
Youngduck Choi\sup{1,2}}
\affiliation{\sup{1}Riiid! AI Research, \sup{2}Yale University, \sup{3}University of Michigan}
\email{
\{yn.lee, dm.shin, hb.loh, jm.lee, piljae, jh.cho, seoyon.park, jh.lee, jineon.baek, byungsoo.kim, youngduck.choi\}@riiid.co}
}

% \author{\authorname{Anonymous Author(s)}
% \affiliation{affiliation}
% \email{email}
% }


% \abstract{The abstract should summarize the contents of the paper and should contain at least 70 and at most 200 words. The text must be set to 9-point font size.}

\abstract{
Student dropout prediction provides an opportunity to improve student engagement, which maximizes the overall effectiveness of learning experiences.
However, researches on student dropout were mainly conducted on school dropout or course dropout, and study session dropout in a mobile learning environment has not been considered thoroughly. 
In this paper, we investigate the study session dropout prediction problem in a mobile learning environment.
First, we define the concept of the study session, study session dropout and study session dropout prediction task in a mobile learning environment.
Based on the definitions, we propose a novel Transformer based model for predicting study session dropout, DAS: \textbf{D}eep \textbf{A}ttentive Study \textbf{S}ession Dropout Prediction in Mobile Learning Environment.
DAS has an encoder-decoder structure which is composed of stacked multi-head attention and point-wise feed-forward networks.
The deep attentive computations in DAS are capable of capturing complex relations among dynamic student interactions.
To the best of our knowledge, this is the first attempt to investigate study session dropout in a mobile learning environment.
Empirical evaluations on a large-scale dataset show that DAS achieves the best performance with a significant improvement in area under the receiver operating characteristic curve compared to baseline models.
}

\onecolumn \maketitle \normalsize \setcounter{footnote}{0} \vfill

\section{Introduction}
Maximizing the learning effect for each individual student is the primary problem in the field of Artificial Intelligence in Education (AIEd).
Prevalent approaches for the problem mainly focus on generating optimal learning path, where an Intelligent Tutoring System (ITS) recommends learning items, such as questions or lectures, with the best efficiency based on student's learning activity records \cite{reddy2017accelerating,zhou2018personalized,Liu:2019:ECS:3292500.3330922}.
However, one should consider not only the efficiency of learning items but also student engagement too, to maximize the overall effectiveness of learning experiences.
Even though the ITS determines the optimal learning path with best efficiency, the educational goal is not achievable if a student drops out of a study session at an early stage.
By predicting student dropout from a study session, an ITS can dynamically modify service strategy to encourage student engagement.

Previously, student dropout research has mainly studied on school dropout \cite{archambault2009student,marquez2016early} and course dropout \cite{liang2016machine,marquez2016early,liang2016big,whitehill2017delving}, where traditional machine learning techniques, such as support vector machines, decision tree, logistic regression, and naive bayes, were commonly used.
With the development of Massive Online Open Courses (MOOC) and the availability of massive user activity data, more complex models based on neural networks were proposed to predict course dropout in the MOOC environment \cite{hansen2019modelling,beres2019sequential,feng2019understanding}.

Unfortunately, despite the active studies conducted on student dropouts, mobile learning environments were not considered thoroughly in the AIEd research community.
Students in a mobile learning environment are prone to be distracted by many external variables, such as phone rings, texting and social applications \cite{harman2011cell,junco2012too,chen2016does}.
As a result, unlike school dropout and course dropout, study session dropout in a mobile learning environment occurs more frequently, which causes shorter study session length. % and is sensitive to student's real-time interactions.
Therefore, directly applying previous approaches for predicting student dropout to study session dropout prediction results in poor performance since it fails to capture the relations of student actions in shorter time frame.

In this paper, we investigate the study session dropout prediction problem in a mobile learning environment.
First, we define the concept of a study session, study session dropout and study session dropout prediction task in a mobile learning environment.
Based on the observation of student interaction data and following the work of \cite{halfaker2015user}, we define a study session as a sequence of learning activities where the time interval between adjacent activities is less than 1 hour.
Accordingly, if a student is inactive for 1 hour, then we define it as a study session dropout.

From the definitions above, we propose a novel Transformer based model for predicting study session dropout, DAS: \textbf{D}eep \textbf{A}ttentive Study \textbf{S}ession Dropout Prediction in Mobile Learning Environment.
DAS consists of an encoder and a decoder that are composed of stacked multi-head attention and point-wise feed-forward networks.
The encoder applies repeated self-attention to the sequential input of question embedding vectors which serve as queries, keys, and values.
The decoder computes self-attention to the sequence of response embedding vectors which are queries, keys and values, and attention with the output of the encoder alternately.
Unlike the original Transformer architecture \cite{vaswani2017attention}, DAS uses a subsequent mask to all multi-head attention networks to ensure that the computation of current dropout probability depends only on the previous questions and responses.
By considering inter-dependencies among entries, and giving more weights to relevant entries for prediction target, the deep attentive computations in DAS are capable of capturing complex relations of student interactions.

We conduct experimental studies on a large-scale dataset collected by an active mobile education application, Santa, which has 21K users, 13M response data points as well as a set of 15K questions gathered since 2016.
We compare DAS with several baseline models and show that it outperforms all other competitors and achieves the best performance with the significant improvement in area under the receiver operating characteristic curve (AUC).

In short, our contributions can be summarized as follows:
\begin{itemize}
    \item We define the problem of study session dropout prediction in a mobile learning environment.
    \item We propose DAS, a novel Transformer based encoder-decoder model for predicting study session dropout, where deep attentive computations effectively capture complex relations among dynamic student interactions.
    \item Empirical studies on a large-scale dataset show that DAS achieves the best performance with a significant improvement in AUC compared to the baseline models.
\end{itemize}

\section{Related Works}
Dropout Prediction is an important problem studied in multiple areas, such as online games \cite{kawale2009churn}, telecommunication \cite{huang2012customer}, and streaming services \cite{chen2018wsdm}.
Predicting dropout in short-term enables dynamic updates of service strategy, which results in longer session lengths of users.
In long-term, it enables the examination of favorable features of the services, from the relations of service features and dropout rates \cite{halawa2014dropout}.

In the field of education, student dropout prediction has been studied in mainly two areas: school dropout \cite{marquez2016early}, and course dropout \cite{liang2016machine}.
The research in \cite{sara2015high} was the first large scale study on high-school dropout.  
This research examined 36,299 students, where the authors state that previous studies were based on a few hundred students.

Compared to previous works on school dropout prediction which are based on relatively small (hundreds of students), massively generated MOOC log data is actively used in the research on course dropout.
These log data include user actions such as user responses, page accesses, registrations, and clickstreams.
For instance, the dataset described in \cite{reich2019mooc} includes data of 12.67 million course registrations from 5.63 million learners. 
Using the large data from MOOC services, machine learning models such as random forest, SVM \cite{lykourentzou2009dropout}, and neural networks \cite{feng2019understanding} are applied to course dropout predictions in education.
The model in \cite{liu2018predicting} based on Long Short Term Memory Networks (LSTM) predicts course dropout in MOOC.
However, there is no research on the prediction of study session dropout in MOOC which happens more frequently than a course dropout. %, though achievable using MOOC log data.

There are works on session dropout in other fields, such as streaming services, medical monitoring \cite{pappada2011neural}, and recommendation services \cite{song2008real}.
A famous problem similar to this topic is the \emph{Spotify Sequential Skip Prediction Challenge}, which is a problem to predict if a user will skip a song given the previous playlist.
The data schema of this problem is similar to the case of predicting study session dropout based on student-question responses.
A major portion of suggested models in the Spotify Challenge are based on neural networks \cite{hansen2019modelling}.

The research in \cite{daroczy2015machine} suggests a model predict session dropout in the LTE network for network optimization purposes.
The research above is based on sequential models such as LSTM, but recently the Transformer is showing higher performance on similar tasks.

The transformer was first introduced in \cite{vaswani2017attention}. It replaced the recurrent layers in the encoder-decoder architecture with multi-head self-attention. The architecture is widely used in natural language processing tasks since the training process can be parallelized. In AIEd, Transformer based model \cite{pandey2019self} shows higher performance than existing seq2seq models \cite{lee2019creating}.

\section{Study Session Dropout in Mobile Learning}

\begin{figure*}[t]
\centering
\includegraphics[width=1.0\textwidth]{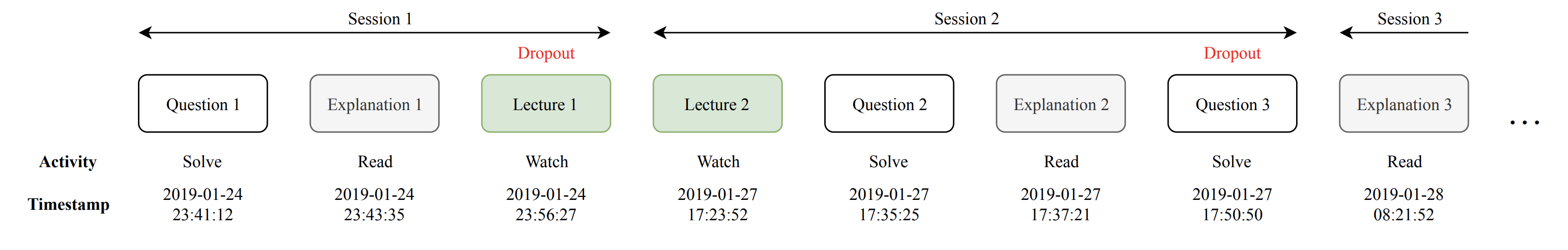}
\caption{Session illustration.}
\label{fig:session}
\end{figure*}

In this section, we define the concept of a study session, study session dropout, and study session dropout prediction task in mobile learning.
A \emph{study session} is a learning process in which the user contiguously participates in learning activities while retaining the educational context of his previous activities.
During each study session, a \emph{study session dropout} happens when the user is inactive in learning for a sufficiently long time, losing the context of his most recent learning process.
Following the work of \cite{halfaker2015user}, which examined user activity data in various fields to conclude that the 1 hour inactivity time gives the best results in clustering user behaviors, we considered inactivity of 1 hour as the threshold for study session dropout.
Note, however, that the criterion for determining study session dropout is flexible and can be chosen according to the particular needs and properties of each ITS.
An example of user activities, sessions, and session dropouts is illustrated in Figure \ref{fig:session}.

\begin{comment}
- which is a sequence of learning activities where each step time is less than 1 hour.
The learning activity can be any of the following: responding to questions, watching videos, and reading educational materials.
Note that the session is not defined by the time between log-in and log-out.
Mobile learning takes place in a multi-tasking environment where the student can simply switch to another task without logging out from the education service.
Users can simply switch screens to send text messages and come back to the service without logging out.
Therefore, using active log-out time as the label is inappropriate in the mobile learning environment.
- If a user is inactive for 1 hour, then we define it as \emph{study session dropout}. Then, the learning activities that follow are considered as a new session. 
- The threshold of 1 hour is based on the research on \cite{halfaker2015user}, which examined user activity data on various fields to conclude that the 1 hour inactivity time gives the best results in regular behaviors for clustering jobs.
- In Figure \ref{fig:session}, we illustrate an example case of user activity, session, and session dropout.
\end{comment}

Identifying study sessions of a mobile ITS user helps the tutoring system to understand student behaviors in coherent units.
For example, one may analyze a student's knowledge state per each activity in a single session to study the short-term learning effect in a session, and per each session to study the long-term learning effects across different sessions.
Along this line, the prediction of study session dropouts may be utilized to guide a student's learning path. 
For instance, students with very short session time may not absorb enough learning materials to maintain what he has learned in his session.
To guide them, an ITS could push pop-up messages that provide educational feedback or encourage additional learning activities to lengthen the study sessions of students. 
For students with very long study sessions, the ITS could suggest the students take a break for more effective learning.
%, or recommend appropriate educational contents 
% By combining these actions, the ITS may motivate students to study in an adequate time span to achieve better learning efficiency by predicting their session dropout behaviors.

However, existing dropout prediction methods cannot be applied to mobile ITSs due to the large difference between traditional and mobile education sessions in length. 
In traditional education, study sessions like school lectures, exercise sessions or timed exams are usually held in environments regulated by instructors.
This enables students to focus solely on given activity in specific space and time limit which results in a longer time span of learning sessions.
In contrast, study sessions in mobile learning do not impose any condition on students' behavior and surrounding environments, allowing students to diverge to other activities. 
According to \cite{chen2016does}, tasks that demand multi-tasking to students like phone rings, texting, and social applications are the main sources of distraction that affect learning activity in a mobile environment.
Another factor for shorter learning time in mobile education is the limitation of the hardware. 
Unlike offline tools like books and blackboards, a mobile device has a smaller screen with constant light emission which is harder to focus for a prolonged time interval, which results in a shorter attention time span.

To this end, we propose the \emph{study session dropout prediction} problem for mobile ITSs.
The task is the prediction of probability that a user drops out from his ongoing study session. 
Unlike offline learning, mobile ITSs can take advantage of automatically collected student behavior data to complete the task in real-time.
For example, a dropout prediction model may utilize the question-response log data of a student as input.
% , including the student’s response, absolute response time, category of each question, and elapsed time, to predict his dropout probability. 
Under this setting, we formalize the problem as the following. Note that utilizing student data aside from question-response logs are also possible, which we defer to future works.

Let 
$$I_1, I_2, \cdots, I_{i} = (e_i, l_i), \cdots, I_{T}$$ 
be the sequence of question-response pairs $I_i$ of a student. 
Here, $e_{i}$ denotes the meta-data of the $i$-th question asked to the student such as question ID or the relative position of the question in the ongoing session.
Likewise, $l_{i}$ denotes the metadata of the student's response to the $i$-th question, which includes the actual response of the student and the time he has spent on the given question. 
The \emph{study session dropout prediction} is the estimation of the probability 
$$P(d_{i} = 1 | I_{1}, \dots, I_{i-1}, e_{i})$$ 
that the student leaves his session after solving the $i$-th question,
where $d_{i}$ is equal to $1$ if the student leaves and $0$ otherwise.

% Let a session be a continuous sequence of learning activities such as responding to questions, reading learning materials, and watching videos. 
% In this paper, we restrict user learning activities to questions only, and train models based on the user responses to the questions.
% Given the history of a user's responses to questions from previous sessions, we predict $d_i$, which is whether the user leaves the session when a certain question is provided.
% In formal language, a user response is a sequence $I_1, \cdots, I_n$ of question-response pair $I_i = (e_i, l_i)$.
% Here $e_i$ denotes the \emph{question information} which is the question meta-data (e.g. id, part, position, relative position in the session) of the $i$'th question asked to the user.
% Note that $e_i$ can differ for all users.
% $l_i$ denotes the meta-data of the user's response (e.g. responded answer, elapsed time, deleted responses) to $e_i$.
% Thus, our model is designed to predict the probability 
% \begin{equation}
%     P(d_i | l_1, \cdots, l_{i-1}, e_1, \cdots, e_{i-1})
% \end{equation}
% that a user leaving the session. 
% Let $d_i \in{\{0,1\}}$ be equal to $1$ if user leaves the session after solving $i$'th question, and $0$ otherwise.

% An education service can provide learning contents by considering both education efficiency and session dropout probability of contents, so that users can study not only efficiently, but also for a longer time to achieve best results.

\section{Proposed Method}

\subsection{Input Representation}

% make table
\begin{table}[b]
    \centering
        \begin{tabular}{cccc}
        \toprule
            Input & Description & $e_i$ & $l_i$ \\
            \midrule
            $id$ &  Question ID & \checkmark & \\ 
            $c$  &  Category & \checkmark & \\ 
            $st$ &  Starting time & \checkmark & \checkmark\\
            $p$  &  Position in input sequence & \checkmark & \checkmark\\
            $sp$ &  Position in session & \checkmark & \checkmark \\
            $r$  &  Response correctness &  & \checkmark\\
            $et$ &  Elapsed time & & \checkmark\\
            $iot$&  Is on time&  & \checkmark\\
            $d$  &  Dropout & & \checkmark\\
        \bottomrule
        \end{tabular}
    \caption{Features of $e_i$ and $l_i$}
    \label{table:feature}
\end{table}

The proposed model predicts student dropout probability based on two feature collections of each interaction $I_i$: the set $e_i$ of question meta-data features and the set $l_i$ of response meta-data features.
The members of $e_i$ and $l_i$ are summarized in Table \ref{table:feature}.
In total, there are a total of nine features constituting $e_i$ and $l_i$:

\begin{itemize}
    \item Question ID $id_i$: The unique ID of each question.
    \item Category $c_i$: Part of the TOEIC exam the question belongs to.
    \item Starting time $st_i$: The time the student first encounters the given problem.
    \item Position in input sequence $p_i$: The relative position of the interaction in the input sequence of our model. Note that this positional embedding was used by  \cite{gehring2017convolutional} to replace the sinusoidal positional encoding introduced in \cite{vaswani2017attention}.
    \item Position in session $sp_i$: The relative position of the interaction in the session it belongs to. The number increments by each problem and resets to 1 whenever a new session starts. 
    \item Response correctness $r_i$: Whether the user's response is correct or not. The value is 1 if correct and 0 otherwise.
    \item Elapsed time $et_i$: The time the user took to respond to given question. 
    \item Is on time $iot_i$: Whether the user responded in the time limit suggested by domain experts. 
    % The explicit time limit we used for each question is denoted in Table \textcolor{blue}{??}.
    The value is 1 if true and 0 otherwise.
    \item Is dropout $d_i$ : Whether the user dropped out after this interaction. The value is 1 if true and 0 otherwise.
\end{itemize}
\begin{comment}
There are a total of 9 features that are embedded for the input of the model:
$id_i$ (the question id), $c_i$ (the category that the question belongs to, which is the TOEIC part number), $s_i$ (the timestamp that the student encounters the problem), $p_i$ (the position number for the model input sequence), $sp_i$ (the position number within the session it belongs to), $r_i$ (the response correctness), $et_i$ (the elapsed time from start time to response), $iot_i$ (whether the user responded in the time limit decided by domain experts), and $d_i$ (label whether the user dropped out when provided that question).

Now we describe some models features in more detail.
The $iot_i$ feature is 
\textcolor{blue}{exactly how is it defined? how many minutes?}
This feature was added since the TOEIC test is a test with time limit.
$sp_i$ is the position number inside the session, which is reset to 1 when a new session starts.
\end{comment}

The vector representations of $e_i$ and $l_i$ are computed by 
summing up the embeddings of aforementioned features.
% The embedded vectors of aforementioned features are added to form $e_i$ and $l_i$.
% Here, $e_i$ and $l_i$ are the vector representations of question feature set and response feature set respectively.
Put formally, we have:
$$e_i = \mathrm{emb}_e(id_i, c_i, st_i, p_i, sp_i)$$
$$l_i = \mathrm{emb}_l(r_{i}, et_{i}, st_{i}, iot_{i}, d_{i}, p_{i}, sp_{i})$$
where $\mathrm{emb}_{-}(.)$ denotes the summation of the corresponding features as distributional vectors. Note that this is equivalent to projecting the concatenation of all features by linearity. 
Separate embeddings are used for features shared across $e_i$ and $l_i$.
For example, the same positional number $sp$ have different distributional vectors in question embedding $\mathrm{emb}_e$ and response embedding $\mathrm{emb}_l$.

\subsection{Model Description}

Our model is based on Transformer, which consists of the encoder and decoder part.
First, we give a brief description of both part.
The encoder consists of $n$ stacked encoder blocks, where each encoder block has two sub-layers, a self-attention layer, and a fully connected feed-forward network layer, where both sub-layers are followed by residual connection and layer normalization.
The encoder takes the sequence of question information embedding $e_1, \dots, e_n$ as the input, where each $e_i$ has dimension $d_\textrm{model}$.
We define $E^i$ as the input of the $i$'th block, and the output of the $i-1$'th block for $i= 2, \dots, n$.
Here, $E^1$ is the input for the first encoder block, which is the output of the embedding layer.
The final output of the encoder is denoted as $h_1, \dots, h_n$

\begin{figure}[t]
\centering
\includegraphics[width=0.35\textwidth]{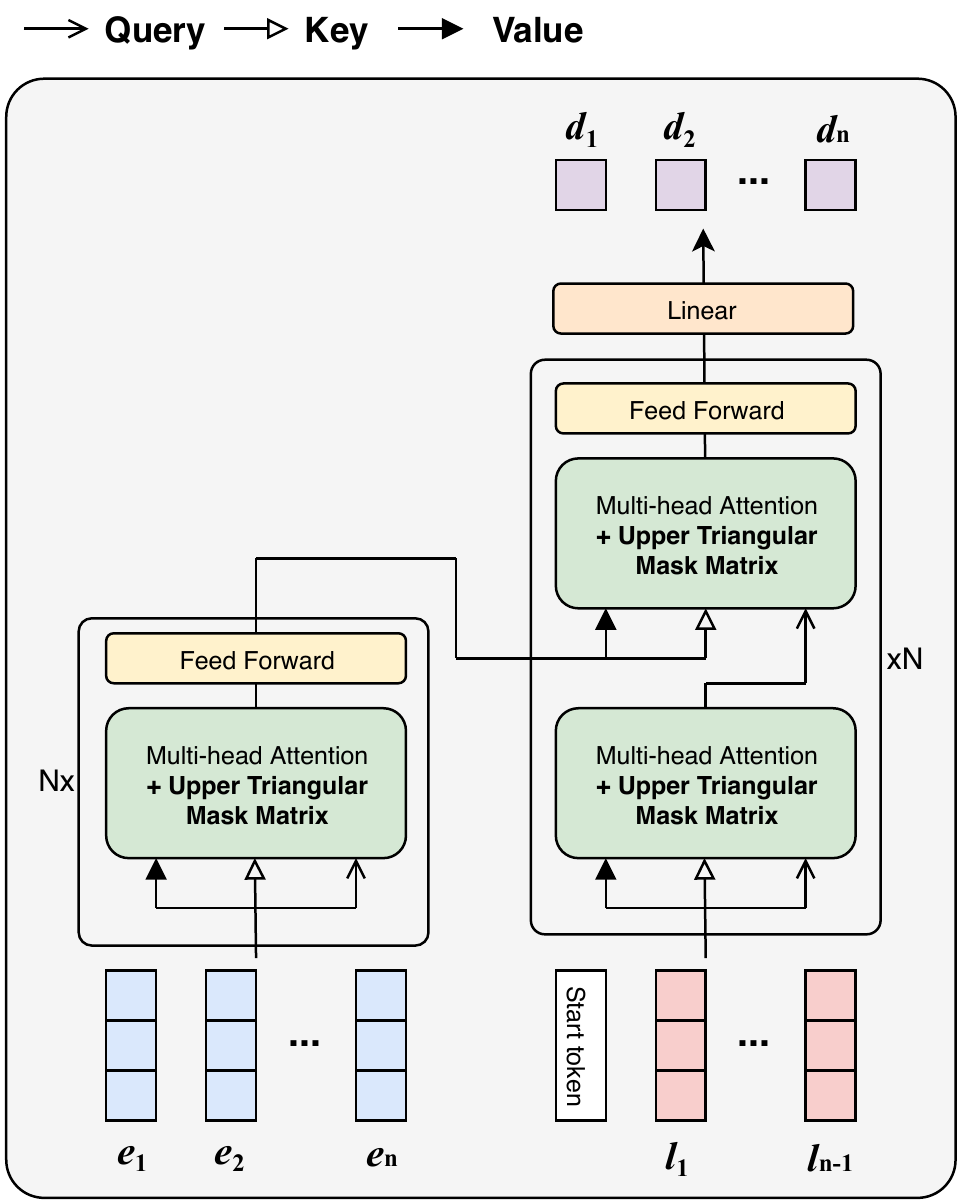}
\caption{The model architecture.}
\label{fig:architecture}
\end{figure}
The decoder also consists of $n$ stacked decoder blocks and one linear layer, where each block has a self-attention layer, an attention layer, and a fully connected feed-forward network layer.
All three sub-layers are followed by residual connection and layer normalization.%Add Norm=> layer?
The decoder takes the user log information embedding $S, l_1, \dots, l_{n-1}$ as input, where $S$ is a constant starting token. 
We define $D^i$ as the input of the $i$'th decoder block, and the output of the $i-1$'th block for $i= 2, \dots, n$.
The additional attention layer takes outputs of self-attention layers as query, and outputs of encoder as key and value. 
% Then, the attention layer takes $h_1, \dots, h_n$ for query, key, and it takes $l_1, \dots, l_{n-1}$ for value.
% The output of the attention layer is fed to a fully connected feed-forward network and 
At the last, we use a learned linear transformation to predict the dropout probabilities $\hat{d}^1. \dots, \hat{d}^n$.
% the final output of the decoder is the predicted probabilities $\hat{d}^1. \dots, \hat{d}^n$ of dropout.

Now, we describe each sub-layer in detail.
The self-attention layer of the $i$'th encoder block takes query, key, value matrices as inputs.
These matrices are given by multiplying $E^i$ to the parameter matrices $W^Q, W^K, W^V$.
\begin{align*}
Q = E^i W^Q & = [q_1, \cdots, q_n]^T \\
K = E^i W^K & = [k_1, \cdots, k_n]^T \\
V = E^i W^V & = [v_1, \cdots, v_n]^T  
\end{align*}
The $i$'th rows $q_i, k_i, v_i$ of $Q, K, V$ are the representations of query vector, key vector, and value vector of the $i$'th input $e_i$.
Let the dimensions of $q_i, k_i, v_i$ be $d_k, d_k, d_v$.
Our model uses multi-head attention, instead of single self-attention layers, to capture various aspects of attention.
Multi-head attention is the method to split the embedding vector into the number of heads $h$ and perform $h$ self-attentions on each part of the split vector.

Then, the attention layer output is computed by multiplying $v_j$ on the Softmax of normalized $q_i \cdot k_j$ for all $j= 1, \dots, n$.
This is written as: 
$$
\textrm{Multihead}(E^i) = \textrm{Concat}(\textrm{head}_1, \cdots, \textrm{head}_h) W^O
$$
Here, each $$\textrm{head}_j=\textrm{Attention}_j(E^i) = \textrm{Softmax} \left( \frac{Q_j K_j^T}{\sqrt{d_k}} \right) V_j,$$
where $Q_j = E^i W^Q_j$, $K_j = E^i W^K_j$ and $V_j = E^i W^V_j$ respectively.
We mask the matrix $QK^T$, where the masking details are described in the following subsection.
%by replacing all the upper-triangular entries $C_{ij}$ of $C$ with $j > i$ to $-\infty$. 
%This process is called subsequent masking, and it prevents the $i$'th attention output from depending on any of the $j$'th entries with $j > i$.

As $M = \textrm{Multihead}(E^i)$ of an attention is a linear combination of values, a position-wise feed forward network is applied to add non-linearity to the model. The formula is given by:
\begin{align*}
F & = (F_1, \cdots, F_n) = \textrm{FFN}(M) \\
 & = \textrm{ReLU}\left(M W^{(1)} + b^{(1)} \right) W^{(2)} + b^{(2)}
\end{align*}
where $W^{(1)}$, $W^{(2)}$, $b^{(1)}$ and $b^{(2)}$ are weight matrices, and bias vectors.

% In summary, the encoder takes the sequence $e_1, \cdots, e_n$ of question information as input and feeds the processed output $h_1, \cdots, h_n$ to the decoder. 

% added the decoder summary part at top and checked until here

% The $i$'th decoder block takes $D^{i}$ as the input and feeds them to a multi-head self-attention layer with masking. 
% Then, a multi-head self-attention layer with masking takes the query and key as $h_1, \dots, h_n$ from the encoder, and the value from the previous layer of the decoder.
% The details for masking are described in the following subsection.
% We finally use a Feed Forward Network to obtain the final predictions $\hat{d}^1, \dots, \hat{d}^n$.
% All three sub-layers are followed by residual connection and layer normalization.

In summary, the whole process of the encoder and decoder can be written by:
\begin{align*}
h_1,\cdots, h_n &= \textrm{Encoder}(e_1, \cdots , e_n)\\
\hat{d}^1, \cdots, \hat{d}^n &= \textrm{Decoder}(S, l_1, \cdots, l_{n-1}, h_1, \cdots ,h_n)
\end{align*}
where $\hat{d}^i$ is the predicted value of 
$$
P( \textrm{Dropout when given } e_i \ \vert \ e_1, \cdots, e_{i-1}, l_1, \cdots, l_{i-1}). $$
% We describe this in detail.

\begin{table*}[t!]
    \label{table:data_structure}
    \centering
    \begin{tabular}{cccccccc}
    \toprule
    \textbf{timestamp}  & \textbf{question\_id} & \textbf{user\_answer} & \textbf{correctness} & \textbf{elapsed\_time} & \textbf{part} & \textbf{session\_id}& \textbf{dropout} \\ \midrule
2019-02-12 09:40:21 & 5279        & c           & 1              & 33        & 5  & 1  & 0        \\ 
2019-02-12 09:40:51 & 5629        & b           & 0              & 26        & 5  & 1  & 0        \\ 
2019-02-12 09:41:10 & 6048        & a           & 1              & 16        & 5  & 1  & 0        \\ 
2019-02-12 09:41:54 & 6158        & b           & 0              & 41        & 2  & 1  & 1        \\ 
2019-02-14 19:32:27 & 5022        & d           & 1              & 30        & 2  & 2  & 0        \\ 
\bottomrule
\end{tabular}
    \caption{User activity log data}
    \label{tab:datatable}
\end{table*}

\subsection{Subsequent Masking}

Transformer models use offset and subsequent masks to handle causality issues.
Appropriate subsequent masks in the sequential data case can let each row represent a time step, so that different time steps can be fed to the network for training.
Here, the details of masking should depend on the context of problem.
A Transformer model for Machine Translation (MT) uses the whole input sentence, and the first $i-1$ words of the partially translated sentence $v_1, \dots, v_{i-1}$ to translate the $i$'th word of the input sentence.
To respect this causality, the sequence $S, V_1, \cdots, V_{i-1}$ with offset by a starting token $S$ enters the decoder from the encoder at step $i$.
We apply this structure of Transformer to session dropout prediction, but with modifications on masking to fit our problem details.

Compared to the original transformer model, our model uses a subsequent mask on all multi-head attention layers (encoder multi-head attention, decoder multi-head attention, encoder-decoder multi-head attention) to prevent invalid attending.
We mask all attention layers to ensure that the computation of $\hat{d}^i$ depends only on the information from the previous questions $e_1, \cdots, e_i$ and responses $l_1, \cdots, l_{i-1}$ on the $i$'th step.
In the MT example, it is natural for the decoder to translate a word by attending all the words before and after the word in the source sentence from the encoder. 
But in our case, attending to future questions $e_{i+1}, \cdots, e_n$ to predict $d_i$ is invalid since further problem suggestions depend on $l_i$. 
Let $I_1, \dots, I_n$ be a sequence of question-response pairs with a student ending the session after solving $e_n$, where the session length is $n$.
Directly applying the MT model to this situation will be predicting each $d_i$ given $e_1, \dots, e_n$ and $l_1, \dots, l_{n-1}$.
However, as described above, our case differs from the MT case $e_j$ for $j > i$ is not given at the point $i$.
Therefore, we apply subsequent masks on future question information to both the input sequence and the in mid-processes.

\section{Dataset}

% \subsection{Data Description}
We use the \emph{Santa} dataset for training our model, which is released in 2019 by mobile AI tutor \emph{Santa} for English education \cite{ednet}.
Specifically, \emph{Santa} aims to prepare students for TOEIC\textsuperscript{\textregistered}(Test Of English for International Communication\textsuperscript{\textregistered}).
The test consists of seven parts divided into listening and reading sessions, with 100 questions assigned for each session.
The final score is subjective and ranges from 0 to 990 in a score gap of 5. 
At the time of writing, the application is available via Android and iOS applications with 1,047,747 users signed up for the service.
In \emph{Santa}, users are provided multiple-choice questions recommended by the \emph{Santa} AI tutor.
After a user responds to a given question, he receives corresponding educational feedback by reading an expert's commentary or watching relevant lecture videos.
Every user responses from 2016 to 2019 are recorded in the \emph{Santa} dataset with the following columns of our interest (see Table \ref{tab:datatable}).

\begin{figure}[b]
\centering
\includegraphics[width=0.5\textwidth]{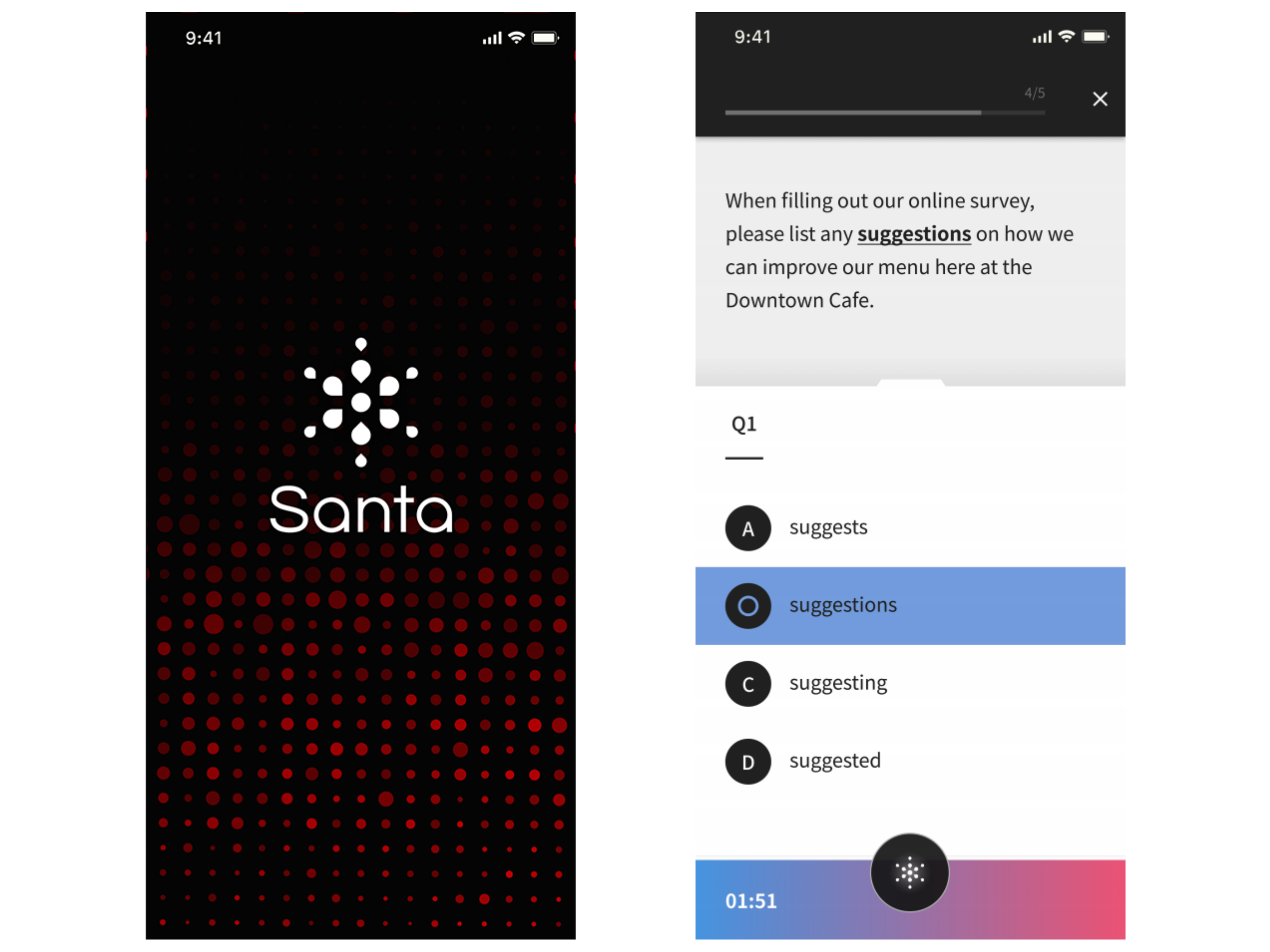}
\caption{User interface of \emph{Santa}.}
\label{fig:ui}
\end{figure}

\begin{itemize}
    \item \texttt{user\_id}: A unique ID assigned to each user. 
    We group the rows having the same \texttt{user\_id} to recover each user's activity log data.
    \item \texttt{timestamp}: The time the user received a question.
    \item \texttt{question\_id}: The ID of the question the user received.
    \item \texttt{user\_answer}: User's response, which is one of the four possible choices A, B, C, or D.
    \item \texttt{correctness}: Correctness of the user's response, which is 1 if correct and 0 otherwise.
    \item \texttt{elapsed\_time}: The time the user took to respond in milliseconds.
    \item \texttt{part}: Part of the TOEIC\textsuperscript{\textregistered} exam the question belongs to, from Part 1 to Part 7.
    % \item \texttt{session_id}: The ID of the session the interaction belongs to, computed by % \texttt{timestamp} with the 1-hour criteria.
    % \item \texttt{is\_dropout}
\end{itemize}
The dataset involves a total of 13,840,169 interactions between 216,575 users and 14,900 questions. 
For training and testing our model, the dataset is split into train set (151,602 users, 9,643,191 responses), validation set (21,658 users, 1,409,323 responses) and test set (43,315 users, 2,787,655 responses) per user basis by the 7:1:2 user count ratio.

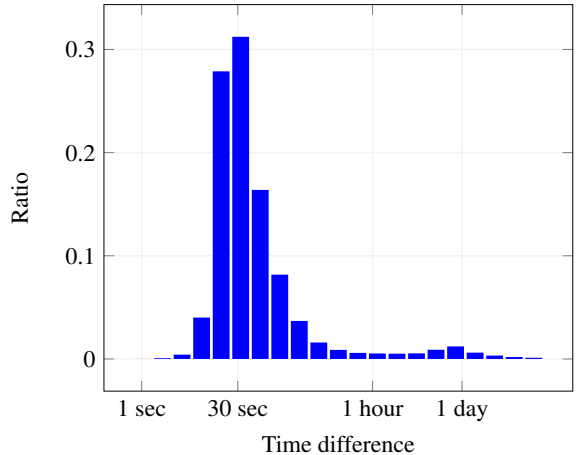
\begin{figure}[t]
\centering
\begin{scaletikzpicturetowidth}{\linewidth}
\begin{tikzpicture}[scale=\tikzscale]
\begin{axis}[
ylabel={Ratio},
xlabel={Time difference},
grid=major,
grid style={line width=.6pt, draw=gray!10},
yticklabels = {0, 0.1, 0.2, 0.3},
xticklabels = {1 sec, 30 sec, 1 hour, 1 day},
ytick={0,0.1,0.2,0.3},
xtick={19.93, 24.85, 31.75, 36.33},
]
\addplot[ycomb,color=blue,line width=7pt]
  coordinates {
(20,0.000037)(21,0.000761)(22,0.004188)(23,0.040137)(24,0.278774)(25,0.312241)(26,0.163863)(27,0.081722)(28,0.036860)(29,0.015962)(30,0.008800)(31,0.005895)(32,0.005316)(33,0.005121)(34,0.005429)(35,0.008922)(36,0.012156)(37,0.006131)(38,0.003276)(39,0.001891)(40,0.001167)
  };
\end{axis}
\end{tikzpicture}
\end{scaletikzpicturetowidth}
\caption{Histogram of time differences between two consecutive interactions of all users in log scale.}
\label{fig:timediff}
\end{figure}

% \subsection{Session Dropout Label}
Following our criterion for identifying session dropout, we divide a user's interactions into different sessions by every inactive time intervals of $\geq 1$ hour.
Figure \ref{fig:timediff} shows that the value of 1 hour separates the time differences of consecutive actions into a large bump with a peak around 30 seconds, and a small bump with a peak around a day.
This suggests that our criterion is reasonable for identifying long intervals between different sessions from short gaps in an ongoing session.

Each separated session is then assigned a unique \texttt{session\_id}.
The last interaction of each session is marked as dropout interaction by setting the value of the column \texttt{dropout} to 1 (the default value of \texttt{dropout} is 0).
Table \ref{tab:datatable} illustrates a sample user's question response data with columns \texttt{session\_id} and \texttt{dropout}.
% Other interactions are marked with 0 for the \texttt{dropout} column. 
The processed data has 772,235 sessions, with an average of 3.57 sessions per user and 17.92 questions per each session.
Among all responses, $1/17.92 = 5.58\%$ are dropout responses.
Each session lasts for 26.00 minutes on average. 

\begin{table}[t]
    \centering
    \begin{tabular}{c|c}
    \hline
    \toprule
     Statistics & \emph{Santa} dataset \\
    \bottomrule
     total user count                   & 216,575 \\
     train user count                   & 151,602 \\
     validation user count              & 21,658 \\
     test user count                    & 43,315 \\
     total response count               & 13,840,169 \\
     train response count               & 9,643,191 \\
     validation response count          & 1,409,323 \\
     test response count                & 2,787,655 \\
     % correct response ratio             & 0.66 \\
     % incorrect response ratio           & 0.34 \\
     % mean interaction sequence length   & 116.21 \\
     % median interaction sequence length & 13 \\
     % max interaction sequence length    & 41,644 \\
    % number of categories                &7 \\
    \bottomrule
    \end{tabular}
    \caption{Statistics of the \emph{Santa} dataset}
    \label{table:statistics}
\end{table}

\section{Experiments}

\subsection{Training Details}
% Since the label of the dataset is imbalanced, we modify the ratio of positive and negative labels to 1:1 by oversampling positive labels. 
% Here, the positive labels are the cases when the user dropped out of the session.
% Since the label $\texttt{dropout}$ is imbalanced,

During training, 
we maintain the ratio of positive and negative $\texttt{dropout}$ labels to 1:1 by over-sampling dropout interactions.
Model parameters that give the best AUC on validation set is chosen for testing.
The best-performing model have $N = 4$ stacked encoder and decoder blocks.
The latent space dimension $d_\textrm{model}$ of query, key, value and the final output of each encoder/decoder block is equal to $512$.
Each multi-head attention layer consists of $h = 8$ heads. 
All model are trained from scratch with weights initialized by Xavier uniform distribution \cite{xavier_unifom}.
% We follow most of the model hyperparameters from \cite{vaswani2017attention} and 
We use the Adam optimizer \cite{kingma2014adam} with hyperparameters $\beta_1$ = 0.9, $\beta_2$ = 0.98, epsilon = $10^{-9}$.
The learning rate $lr = d_\textrm{model}^{-0.5} \cdot \min (step\_num^{-0.5}, step\_num \cdot ws^{-1.5})$ follows that of \cite{vaswani2017attention} with $ws = 6000$ warmup steps.
A dropout ratio of 0.5 was applied during training.
% We choose the model parameters that give the best result on validation set and  evaluate the model with the test set.

\subsection{Experiment Results}

We compare DAS with current state-of-the-art session dropout models based on Long Short-Term Memory (LSTM) and Gated Recurrent Unit (GRU) \cite{hussain2019using} by training and testing the models with \emph{Santa} dataset. 
As LSTM and GRU models take only one input for each index, 
we set the $i$'th input of the models as $in_{i-1} + e_i$,
where $in_{i-1} = e_{i - 1} + l_{i-1}$ is the representation of the previous ($(i-1)$'th) question-response interaction. For the first input, we replace $in_{i-1}$ with a starting token. 
Like DAS, optimal parameters for LSTM and GRU were found by maximizing the AUC over validation set.
All models were trained and tested with sequence size 5 and 25 to find the best input length for session dropout prediction. 
The results in Table \ref{table:comparison} shows that DAS with sequence size 5 outperforms best LSTM and GRU models by 12.2 points.
% In the setting of input sequence size 5, the AUC of our model with dropout rate 0.5 is higher than LSTM by 7.6 points and GRU by 10.4 points respectively.

% These models also capture the sequential properties of the data but do not capture the question features for embedding as the proposed model based on Transformer. 
% Area Under Curve (AUC) is used for the performance metrics. 
% AUC shows the true positive rate against the false-positive rate at various thresholds. 
% True positive rate is the ratio of the number of samples correctly predicted as positive to the total number of real positive samples, and the false-positive rate is the ratio of the number of samples incorrectly predicted as positive to the total number of real negative samples.

\begin{table}[!b]
\centering
\begin{tabular}{llll}
\toprule
Methods & AUC \\
\midrule
LSTM-25 & 0.5786 & \\ % old: 0.5958
LSTM-5 & 0.6830 & \\ % old: 0.6852
\midrule
GRU-25 & 0.5640 & \\ % old: 0.5878
GRU-5 & 0.6830 & \\ % old: 0.6681
\midrule
DAS-25 & 0.6895 &  \\ % old: 0.6856
DAS-5 & \textbf{0.7661} & \\ % old: 0.7379
\bottomrule
\end{tabular}
\caption{Model comparison}
\label{table:comparison}
\end{table}

\subsection{Ablation study}
In this section, we present our ablation studies on different sequence sizes and input feature combinations of DAS. 

\begin{figure}[!t]
\centering
\begin{tikzpicture}
\begin{axis}[
width = 8.5cm,
xlabel={\# of epochs},
every tick label/.append style={scale=0.7},
% label style={scale=0.8},
    y tick label style={
        /pgf/number format/.cd,
            fixed,
            fixed zerofill,
            precision=2,
        /tikz/.cd
    },
ytick={0.66, 0.68, 0.70, 0.72, 0.74, 0.76},
legend style={legend pos=south east, legend cell align=left, font=\fontsize{6}{10}\selectfont}
]
\addplot[color=red, mark=asterisk ]
  coordinates {
%(1, 0.7078)(2, 0.7127)(3, 0.7129)(4, 0.7132)(5, 0.7132)(6, 0.7126)(7, 0.7111)(8, 0.7106)(9, 0.709)(10, 0.7081)
(1, 0.7112)(2, 0.7152)(3, 0.7163)(4, 0.7169)(5, 0.716)(6, 0.7172)(7, 0.7155)(8, 0.7143)(9, 0.7128)(10, 0.7113)
  };
  \addlegendentry{2};
\addplot[color=blue, mark=triangle]
  coordinates {
%(1, 0.728)(2, 0.7317)(3, 0.7337)(4, 0.7345)(5, 0.7357)(6, 0.7354)(7, 0.736)(8, 0.7355)(9, 0.7362)(10, 0.737)
(1, 0.7315)(2, 0.7379)(3, 0.7449)(4, 0.7552)(5, 0.7583)(6, 0.761)(7, 0.7624)(8, 0.7638)(9, 0.7642)(10, 0.7661)
  };
  \addlegendentry{5},

\addplot[color=black, mark=+]
  coordinates {
%(1, 0.7141)(2, 0.7215)(3, 0.7241)(4, 0.7256)(5, 0.7316)(6, 0.7266)(7, 0.7297)(8, 0.7293)(9, 0.7326)(10, 0.7319)
(1, 0.717)(2, 0.7176)(3, 0.7248)(4, 0.7257)(5, 0.7308)(6, 0.7332)(7, 0.7351)(8, 0.7338)(9, 0.7372)(10, 0.739)
  };
  \addlegendentry{8};
 
  \addplot[color=cyan, mark=diamond]
  coordinates {
%(1, 0.6861)(2, 0.7065)(3, 0.7078)(4, 0.711)(5, 0.7176)(6, 0.7162)(7, 0.7159)(8, 0.7214)(9, 0.7223)(10, 0.7235)
(1, 0.6924)(2, 0.7041)(3, 0.7167)(4, 0.7139)(5, 0.7159)(6, 0.7232)(7, 0.727)(8, 0.7282)(9, 0.7329)(10, 0.7388)
  };
  \addlegendentry{10};
  \addplot[color=magenta, mark=star]
  coordinates {
%(1, 0.6559)(2, 0.6696)(3, 0.6744)(4, 0.6806)(5, 0.6846)(6, 0.6809)(7, 0.6842)(8, 0.6845)(9, 0.6853)(10, 0.6829)
(1, 0.6709)(2, 0.6754)(3, 0.676)(4, 0.6894)(5, 0.6916)(6, 0.6967)(7, 0.698)(8, 0.6959)(9, 0.6894)(10, 0.6895)
  };
  \addlegendentry{25};
\end{axis}
\end{tikzpicture}
\caption{AUC of DAS models with different sequence sizes.}
\label{fig:seq_size}
\end{figure}
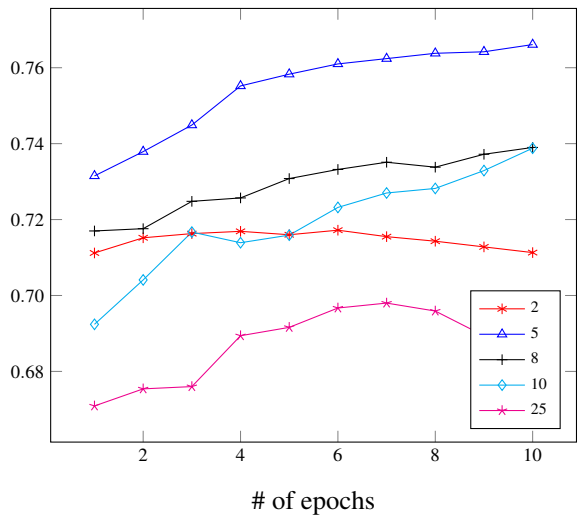

% \subsubsection{Sequence Size}
First, we run an ablation study on different input sequence sizes of DAS.
The results in Table \ref{table:seq_size} show that a sequence size of 5 produces the best AUC. 
This trend is consistent across all epochs (see Figure \ref{fig:seq_size}), suggesting that session dropouts are more correlated with latest student interactions than earlier ones. 
However, as the sequence size of 2 overfits quickly with poor results, it is also observed that the model needs a sufficient amount of context for effective prediction.

% We run an ablation study on different sequence sizes on the model. Table \ref{table:seq_size} shows that adequately short size of an input sequence gives the better result. The model with input sequence size 5 shows the best result. There is 7.6 points of AUC improvement, compared to the model of input sequence size 25. But the model performance with extremely reduced sequence size, such as size 2, is worse than the model with the input sequence size of longer than 5. 

% \begin{figure}[h!]
% \centering
% \includegraphics[width=0.45\textwidth]{fig/seq_size_auc.png}
% \caption{AUC}
% \label{fig:architecture}
% \end{figure}

% \subsubsection{Input Features}

\begin{table}[!tb]
\centering
\begin{tabular}{ccc}
\toprule
Sequence Size & AUC \\
\midrule
2 & 0.7172  \\
5 & \textbf{0.7661} \\
8 & 0.7390 \\
10 & 0.7388 \\
25 & 0.6980 \\
\bottomrule
\end{tabular}
\caption{Effects of sequence size on AUC}
\label{table:seq_size}
\end{table}

% \begin{figure}[h!]
% \centering
% \includegraphics[width=0.45\textwidth]{fig/input_variation_auc.png}
% \caption{AUC}
% \label{fig:architecture}
% \end{figure}

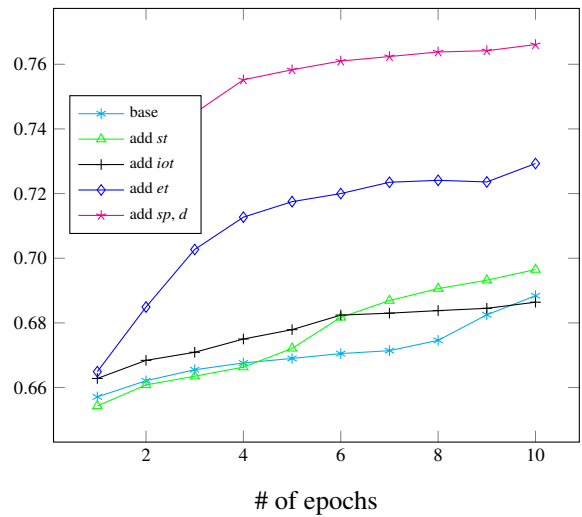
\begin{figure}[t]
\centering
\begin{tikzpicture}
\begin{axis}[
width = 8.5cm,
xlabel={\# of epochs},
every tick label/.append style={scale=0.7},
% label style={scale=0.8},
    y tick label style={
        /pgf/number format/.cd,
            fixed,
            fixed zerofill,
            precision=2,
        /tikz/.cd
    },
ytick={0.66, 0.68, 0.70, 0.72, 0.74, 0.76},
xtick={2, 4, 6, 8, 10},
legend style={legend pos=north west, at={(0.03, 0.8)}, legend cell align=left, font=\fontsize{6}{10}\selectfont}
]
\addplot[color=cyan, mark=asterisk ]
   coordinates {
%(1, 0.6575)(2, 0.6624)(3, 0.6632)(4, 0.6636)(5, 0.6657)(6, 0.6668)(7, 0.6681)(8, 0.6688)(9, 0.6705)(10, 0.6715)
(1, 0.6571)(2, 0.6621)(3, 0.6655)(4, 0.6676)(5, 0.669)(6, 0.6705)(7, 0.6714)(8, 0.6746)(9, 0.6825)(10, 0.6884)

   };
  \addlegendentry{base};
\addplot[color=green, mark=triangle]
   coordinates {
%(1, 0.647)(2, 0.6525)(3, 0.6571)(4, 0.6574)(5, 0.6596)(6, 0.6589)(7, 0.6612)(8, 0.6615)(9, 0.6626)(10, 0.6627)
(1, 0.6543)(2, 0.6608)(3, 0.6635)(4, 0.6663)(5, 0.6721)(6, 0.6817)(7, 0.6869)(8, 0.6906)(9, 0.6932)(10, 0.6965)

   };
  \addlegendentry{add \textit{st}},

\addplot[color=black, mark=+]
   coordinates {
%(1, 0.6612)(2, 0.6673)(3, 0.6703)(4, 0.6724)(5, 0.6747)(6, 0.6763)(7, 0.6773)(8, 0.6784)(9, 0.6798)(10, 0.6796)
(1, 0.6628)(2, 0.6684)(3, 0.6709)(4, 0.675)(5, 0.6779)(6, 0.6824)(7, 0.683)(8, 0.6838)(9, 0.6845)(10, 0.6864)

   };
  \addlegendentry{add \textit{iot}};
 
  \addplot[color=blue, mark=diamond]
   coordinates {
%(1, 0.6607)(2, 0.6677)(3, 0.6701)(4, 0.6746)(5, 0.6806)(6, 0.6827)(7, 0.6872)(8, 0.6912)(9, 0.6954)(10, 0.6987)
(1, 0.6649)(2, 0.6849)(3, 0.7027)(4, 0.7127)(5, 0.7175)(6, 0.72)(7, 0.7235)(8, 0.7241)(9, 0.7236)(10, 0.7293)

   };
  \addlegendentry{add \textit{et}};
  \addplot[color=magenta, mark=star]
   coordinates {
%(1, 0.728)(2, 0.7317)(3, 0.7337)(4, 0.7345)(5, 0.7357)(6, 0.7354)(7, 0.736)(8, 0.7355)(9, 0.7362)(10, 0.737)
(1, 0.7315)(2, 0.7379)(3, 0.7449)(4, 0.7552)(5, 0.7583)(6, 0.761)(7, 0.7624)(8, 0.7638)(9, 0.7642)(10, 0.7661)

   };
  \addlegendentry{add \textit{sp}, \textit{d}};
\end{axis}
\end{tikzpicture}
\caption{AUC of DAS models with different input feature combinations.}
\label{fig:timestep}
\end{figure}

\begin{table}[t]
\begin{adjustbox}{width=0.47\textwidth}
\centering
\begin{tabular}{lllll}
\hline
\toprule 
Name & Encoder Inputs & Decoder Inputs & AUC  \\

\midrule
Base&$id, c, p $ & $r, p$ & 0.6884 \\ 
\midrule
add $st$&$id, c, p, st $ & $r, p, st$ & 0.6965 \\ 
add $iot$ & $id, c, p, st $ & $r, p, st, iot $ & 0.6864 \\
add $et$ & $id, c, p, st $ & $r, p, st, iot, et$ & 0.7293 \\ 
\midrule
add $sp, d$ & $id, c, p, st, sp$ & $r, p, st, iot, et, sp, d $ & \textbf{0.7661} \\
\bottomrule

\end{tabular}
\end{adjustbox}
\caption{Effects of input features on AUC}
\label{table:input}

\end{table}

% Table \ref{table:input} shows how the AUC, and Recall changes by using more input features.
% We use question id $id$, TOEIC part $c$, and input sequence position $p$ for encoder inputs.
% The input sequence position $p$ is irrelevant to the question information itself, but is required for the model input.
% For the decoder, we use response $r$, and input sequence position $p$.
% Metrics decrease by adding start time $st$ to the input feature, but metrics improve by adding more features one-by-one: whether the response was in time limit (is on time) $iot$, elapsed time $et$, session position $sp$, dropout $d$.
% Here, the session position is the position number in each session of the user, which ranges from 1 to the session length.

% The most significant feature is session position $sp$, which shows the greatest improvement of metrics when added to the input features.
% This shows that session dropouts is strongly associated with the number of solved questions.

Second, we run an ablation study on different combinations of input features for DAS.
Table \ref{table:input} shows that test AUC gradually increases as we include input features one by one, achieving the highest value with all the features in Table \ref{table:feature}.
From Figure \ref{fig:timestep}, it can be seen that elapsed time $et$, session position $sp$ and session dropout labels of previous interactions $d$ improves the overall AUC prominently. 
The large improvement with elapsed time $et$ suggests that the time a student spends on learning activity is highly correlated to his dropout probability.  
Also, as it is likely for students with sufficient amount of learning activities to drop his session, it is natural for the features $sp$ and $d$ to be effective for student dropout prediction task.

\section{Conclusion}
Student dropout prediction provides an opportunity to improve student engagement and maximize the overall effectiveness of learning experience.
% However, existing works on study dropout were mainly conduced on school and course dropout 
However, student dropout research has been mainly conducted on school dropout and course dropout, and study session dropouts in mobile learning environments were not considered thoroughly in literature. % in artificial intelligence in education research community.
In this paper, we defined the problem of study session dropout prediction in a mobile learning environment.
We proposed DAS, a novel Transformer based encoder-decoder model for predicting study session dropout, in which the deep attentive computations effectively capture the complex relations among dynamic student interactions.
Empirical studies on a large-scale dataset showed that DAS achieves the best performance with a significant improvement in AUC compared to the baseline models.

\bibliographystyle{apalike}
{\small
\bibliography{ref.bib}}

\end{document}